\documentclass[11pt,letterpaper]{article}

\usepackage[margin=1in]{geometry}
\usepackage[utf8]{inputenc}
\usepackage[T1]{fontenc}
\usepackage[english]{babel}
\usepackage{lmodern}
\usepackage{amsmath,amssymb,bm}
\usepackage{graphicx}
\usepackage{array,booktabs}
\usepackage{caption,subcaption}
\usepackage{enumitem}
\usepackage{listings}
\usepackage{microtype}
\usepackage[hidelinks]{hyperref}

\hypersetup{
  pdftitle={Beyond Affine Transformations: A Soft Dominance Layer for Coordinate-Wise Neural Computation},
  pdfauthor={Mariano Rivera}
}
\setlist[itemize]{itemsep=0pt,parsep=0pt}
\newcommand{\keywords}[1]{%
  \par\medskip\noindent{\small\textbf{Keywords:} #1}\par\medskip
}

\title{Beyond Affine Transformations: A Soft Dominance Layer for Coordinate-Wise Neural Computation}
\author{Mariano Rivera\\
  {\small Centro de Investigaci\'on en Matem\'aticas, A.C.}\\
  {\small Guanajuato, 36023, M\'exico}\\
  {\small\texttt{mrivera@cimat.mx}}}
\date{September 10, 2026}

\begin{document}
\maketitle

\begin{abstract}
This paper presents a preliminary study of an alternative to the affine transformation underlying conventional neural-network layers. In the proposed \emph{Soft Dominance Layer}, each output unit compares input coordinates with a learnable reference vector and aggregates smooth inequality responses. A sigmoid relaxation makes the comparisons differentiable, while a sharpness parameter $\alpha$ controls their transition toward hard threshold decisions. The aim is to examine the trainability and direct threshold interpretation of this primitive, not to claim a replacement for affine layers. In single-run MNIST experiments, the highest observed Soft Dominance accuracy is $0.9061$ without annealing and $0.9173$ with annealing, compared with $0.9827$ for the MLP baseline. These descriptive results do not establish reliable configuration rankings or a statistically supported annealing benefit. Learned reference vectors exhibit spatial structure, providing qualitative evidence of structured learning. Repeated-seed experiments and broader datasets are required to assess robustness and practical relevance beyond this proof of concept.
\end{abstract}

\section{Introduction}

Despite the diversity of modern neural-network architectures, a large fraction of their computation is still built from a recurring primitive: an affine transformation followed by a nonlinear activation. For an input vector $\bm x$, a conventional dense layer computes
\begin{equation}
    \bm y = \bm W\bm x + \bm b,
    \label{eq:affine_mapping}
\end{equation}
possibly followed by a nonlinear function. Variants of this pattern appear in multilayer perceptrons (MLPs)~\cite{rumelhart1986backprop}, convolutional networks~\cite{lecun1998cnn}, and the position-wise feed-forward blocks of Transformer architectures~\cite{vaswani2017attention}. The success of these models has made matrix multiplication one of the dominant operations in contemporary machine learning hardware and software~\cite{nvidiaTensorCores}.

The prevalence of Eq.~\eqref{eq:affine_mapping}, however, does not imply that affine mappings are the only useful primitive from which trainable networks can be constructed. Previous work has explored binary and low-precision computation~\cite{hubara2016binarized}, sparsity~\cite{han2015connections}, low-rank structure~\cite{denton2014linear}, differentiable logic, and alternative algebras. For example, Min--Max--Plus networks combine conventional linear operations with min-plus and max-plus mappings inspired by tropical arithmetic~\cite{luo2021minmaxplus}. These directions motivate a broader question: \emph{what other simple, differentiable computational rules can serve as neural layers?}

This paper investigates one deliberately simple answer to that question. We replace the dot-product aggregation of a conventional neuron by independent coordinate-wise comparisons between the input and a learnable reference vector. A smooth sigmoid approximation makes each comparison differentiable, and summation produces a soft count of the satisfied inequalities. We call the resulting operator the \emph{Soft Dominance Layer}.

The motivation is primarily structural rather than computational. In a conventional neuron, the parameters of a row of $\bm W$ define a projection direction. In Soft Dominance, the corresponding parameters define coordinate-wise thresholds. This gives the layer a different geometry and a direct interpretation in terms of learned reference values. The present sigmoid implementation is not expected to be faster than highly optimized matrix multiplication; any computational advantage would require approximations, dedicated kernels, or specialized hardware. Our immediate objective is more modest: to determine whether such a non-affine, comparison-based layer can be optimized by standard gradient methods and can learn meaningful internal structure.

Accordingly, this paper focuses only on the simplest additive Soft Dominance formulation. Richer comparison directions, alternative aggregation rules, and more expressive dominance-based basis functions are intentionally left outside the scope of this contribution and reserved for future work.

The contributions are threefold. First, we formulate hard and differentiable coordinate-wise dominance counts and their normalized interpretation. Second, we describe a constrained threshold parameterization and a sharpness-annealing strategy for training the layer. Third, we provide a proof-of-concept evaluation on MNIST against a conventional MLP and examine the organization of the learned threshold vectors. The goal of the experiment is to establish feasibility and expose limitations rather than to claim predictive superiority over affine layers.

\section{Related Work}

The Soft Dominance formulation intersects several lines of work, but its individual mathematical ingredients---smooth thresholds, logistic relaxations, and additive aggregation---have clear precedents. We therefore position the present contribution as a particular \emph{layer organization and interpretation}, rather than as the introduction of a new sigmoid or a new generic differentiable-counting mechanism.

\subsection{Alternative neural computation}

A broad family of methods seeks computational primitives other than unrestricted floating-point affine mappings. Binary networks replace conventional weights and activations by low-precision representations~\cite{hubara2016binarized}, while pruning and low-rank decompositions alter the structure of standard layers~\cite{han2015connections,denton2014linear}. More directly related to the present motivation, Min--Max--Plus networks use min-plus and max-plus operations as trainable mappings~\cite{luo2021minmaxplus}. Soft Dominance follows the same broad philosophy of exploring a different primitive, but its basic operation is a coordinate-wise comparison against a learned reference value.

\subsection{Differentiable comparisons and soft counts}

Smooth approximations to discrete comparisons are well established. Qin, Liu, and Li~\cite{qin2010ranking}, for example, approximate non-differentiable ranking quantities using logistic functions of pairwise score differences; the resulting expressions include sums of smooth comparisons and therefore provide a direct precedent for differentiable counting by logistic relaxation. Differentiable decision-tree models similarly replace hard routing decisions by soft gates so that thresholds can be optimized by gradient methods~\cite{kontschieder2015forests}.

Recent differentiable logic networks also employ trainable smooth threshold units as components of larger logical architectures~\cite{yue2024dln}. These works show that the elementary comparison
\begin{equation}
    \sigma\!\left(\alpha(W_{ij}-x_j)\right)
\end{equation}
is not novel by itself. The distinction in Soft Dominance is that a complete row $\bm W_i$ is treated as a learnable reference vector and all coordinate-wise comparison responses are aggregated directly into a single neural-unit output.

\subsection{Order, monotonicity, and additive structure}

Vendrov et al.~\cite{vendrov2016order} introduced \emph{order embeddings} for representing hierarchical relations between images and language.
Their framework explicitly models a partial order in an embedding space.
In coordinate-wise form, an order relation can be expressed conceptually as
\begin{equation}
    \bm x \preceq \bm y
    \quad \Longleftrightarrow \quad
    x_j \le y_j,
    \qquad \forall j.
\end{equation}
The central objective of that work is to learn representations that respect hierarchical relationships, rather than to construct a neural layer that counts or probabilistically aggregates coordinate-wise threshold comparisons.

Monotonic neural networks~\cite{sill1997monotonic}, Deep Lattice Networks~\cite{you2017deep}, and Hierarchical Lattice Layers~\cite{yanagisawa2022hierarchical} also incorporate order constraints into trainable architectures. Their computational mechanisms differ, however, from the direct comparison-and-sum operator considered here.

A separate connection concerns the additive structure of the operator. Neural Additive Models (NAMs) represent predictions as sums of learned univariate feature functions~\cite{agarwal2021nam}. Each Soft Dominance unit likewise sums univariate contributions, but restricts each contribution to a monotone sigmoid centered at a learned threshold. Thus, the connection to order-based models concerns the coordinate-wise comparisons, whereas the connection to additive models concerns their aggregation. This distinction clarifies both the interpretation and the expressive restrictions of an individual Soft Dominance unit.

\subsection{Positioning of this work}

The goal of this paper is therefore not to claim that sigmoid thresholding or smooth counting is new. Rather, we investigate the specific use of
\begin{equation}
    \sum_j \sigma\!\left(\alpha(W_{ij}-x_j)\right)
\end{equation}
as a neural-layer primitive whose parameters are interpreted collectively as a learned reference point in a coordinate-wise ordered space. The MNIST study is intended as an initial feasibility experiment for this alternative layer design. A broader treatment of more expressive dominance-based constructions is left for subsequent work.

\section{Soft Dominance Layer}

\subsection{Hard dominance count}

For an input $\bm x\in[-1,1]^d$ and an output unit $i$, define
\begin{equation}
    C_i(\bm x)
    =
    \sum_{j=1}^{d}
    \mathbf{1}\!\left[x_j<W_{ij}\right].
\end{equation}

The value satisfies
\begin{equation}
    C_i(\bm x)\in\{0,1,\ldots,d\}.
\end{equation}

Thus, $C_i(\bm x)$ represents the number of coordinates in which $\bm W_i$ dominates $\bm x$.

A normalized version is
\begin{equation}
    D_i(\bm x)
    =
    \frac{1}{d}
    \sum_{j=1}^{d}
    \mathbf{1}\!\left[x_j<W_{ij}\right],
\end{equation}
with
\begin{equation}
    D_i(\bm x)\in[0,1].
\end{equation}
This quantity can be interpreted as the fraction of dominated coordinates.

\subsection{Differentiable relaxation}

We replace the indicator function by
\begin{equation}
    p_{ij}
    =
    \sigma\!\left(
        \alpha(W_{ij}-x_j)
    \right).
\end{equation}

The value $p_{ij}$ is a smooth approximation to the binary event $x_j<W_{ij}$.
For large $\alpha$,
\begin{equation}
    p_{ij}
    \approx
    \mathbf{1}\!\left[x_j<W_{ij}\right].
\end{equation}

The unnormalized Soft Dominance operator is then
\begin{equation}
    S_i(\bm x)
    =
    \sum_{j=1}^{d}
    \sigma\!\left(
        \alpha(W_{ij}-x_j)
    \right).
\end{equation}

Including a learnable bias yields
\begin{equation}
    \boxed{
    y_i
    =
    \sum_{j=1}^{d}
    \sigma\!\left(
        \alpha(W_{ij}-x_j)
    \right)
    +b_i
    }
    \label{eq:softdominance}
\end{equation}

A normalized alternative is
\begin{equation}
    \boxed{
    y_i
    =
    \frac{1}{d}
    \sum_{j=1}^{d}
    \sigma\!\left(
        \alpha(W_{ij}-x_j)
    \right)
    +b_i.
    }
\end{equation}

Without the bias, the normalized response lies in $[0,1]$ and has a particularly simple interpretation as a \emph{soft dominance score}.

\subsection{Geometric interpretation}

A conventional neuron
\begin{equation}
    y_i=\bm w_i^\top \bm x+b_i
\end{equation}
first projects the full input onto a single direction.
Its basic geometry is therefore associated with hyperplanes.

In contrast, a Soft Dominance unit performs $d$ separate comparisons:
\begin{equation}
    x_j<W_{ij},
    \qquad j=1,\ldots,d.
\end{equation}
The vector $\bm W_i$ acts as a multidimensional threshold or reference point.

In the hard-comparison limit, complete dominance corresponds to
\begin{equation}
    \bm x \prec \bm W_i,
\end{equation}
that is,
\begin{equation}
    x_j<W_{ij},
    \qquad \forall j.
\end{equation}

The soft-count formulation provides a graded notion of dominance even when only a subset of coordinates satisfies the ordering relation.

\subsection{Constraining the thresholds}

Since the input is assumed to satisfy
\begin{equation}
    \bm x\in[-1,1]^d,
\end{equation}
it is natural to constrain the learned thresholds to the same range.

Instead of optimizing $\bm W$ directly, an unconstrained parameter matrix $\bm\Theta$ can be introduced:
\begin{equation}
    W_{ij}
    =
    \tanh(\Theta_{ij}).
\end{equation}

This guarantees
\begin{equation}
    W_{ij}\in(-1,1)
\end{equation}
while preserving differentiability.

\subsection{Computational and optimization perspective}

For $d$ inputs and $m$ outputs, Soft Dominance uses $md$ threshold parameters, the same order of parameter count as a dense affine layer. Its forward pass replaces scalar products by $md$ coordinate differences, sigmoid evaluations, and additions. Consequently, the present implementation should not be interpreted as a computationally cheaper substitute for a highly optimized matrix multiplication. The interest here is the change in computational structure and inductive bias. In the hard-comparison limit, or under suitable low-cost approximations to the sigmoid, the operator may also motivate compare-and-accumulate implementations that deserve separate investigation.

For the normalized bias-free response
\begin{equation}
    y_i=\frac{1}{d}\sum_j p_{ij},
    \qquad
    p_{ij}=\sigma\!\left(\alpha(W_{ij}-x_j)\right),
\end{equation}
we have
\begin{equation}
    \frac{\partial y_i}{\partial x_j}
    =-\frac{\alpha}{d}p_{ij}(1-p_{ij})\le 0.
\end{equation}
Thus, each Soft Dominance unit is individually non-increasing in every input coordinate. This property gives the layer a clear order-based bias, but also restricts the interactions that a single unit can represent. The sharpness parameter $\alpha$ controls a trade-off: small values provide broad transition regions and useful gradients, whereas large values more closely approximate discrete comparisons but can increase sigmoid saturation.

\subsection{Trainable and fixed parameters}

The layer can support the following configurations:
\begin{itemize}
    \item trainable $\bm W$ and trainable $\bm b$;
    \item trainable $\bm W$ and fixed $\bm b$;
    \item fixed $\bm W$ and trainable $\bm b$;
    \item fixed $\bm W$ with no bias.
\end{itemize}

This flexibility allows the layer to represent either fully learned threshold prototypes or predetermined order templates.

\section{Proof-of-Concept Evaluation}

We use the MNIST handwritten-digit dataset~\cite{lecun1998cnn} as a proof-of-concept classification problem. The purpose of this experiment is not to establish state-of-the-art performance, but to test whether the proposed non-affine hidden layer can be trained reliably and to identify its current limitations. We compare it with a conventional multilayer perceptron (MLP) whose hidden block consists of a linear layer, dropout~\cite{srivastava2014dropout} with a rate of $0.2$, and a rectified linear unit (ReLU)~\cite{nair2010relu}. In the proposed architecture, the linear--ReLU hidden block is replaced by a Soft Dominance Layer followed by dropout. Both architectures are evaluated with 64, 128, 256, and 512 hidden units. This is a width-matched architectural comparison rather than a parameter- or compute-matched benchmark.

All models were trained for 100 epochs using the Adam optimizer with a learning rate of $1\times10^{-3}$ and a batch size of 32. The MNIST pixel intensities, initially scaled to $[0,1]$, were normalized using a mean of $0.5$ and a standard deviation of $0.5$, mapping them to $[-1,1]$. For the Soft Dominance models, the initial comparison sharpness was set to $\alpha_0=5$.

We additionally evaluate a sharpness-annealing schedule for the comparison parameter $\alpha$. The parameter remains at $\alpha_0$ throughout the first 20 training epochs. It is multiplied by $1.3$ immediately after epoch 20 and after every 10 epochs thereafter (epochs 30, 40, and so on). With epochs indexed from one, let $\alpha_t$ denote the value after completing epoch $t$ and applying any scheduled update:
\begin{equation}
    \alpha_t
    =
    \alpha_0\,1.3^{\max\left\{0,1+\left\lfloor (t-20)/10\right\rfloor\right\}}.
    \label{eq:alpha-annealing}
\end{equation}
The updated value is used in the following epoch. Thus, optimization begins with relatively smooth comparisons and progressively sharpens the sigmoid approximation. Table~\ref{table:results} summarizes the classification results. The suffix ``Ann.'' denotes models trained with the schedule in Eq.~\eqref{eq:alpha-annealing}.

\begin{table}[ht]
\centering
\caption{Single-run MNIST classification results. Each row reports one run; variability across random seeds was not measured. ``Ann.'' denotes $\alpha$ annealing.}
\label{table:results}
\begin{tabular}{lrrr}
\toprule
Method & Accuracy & Recall & Precision \\
\midrule
MLP (64) & 0.9748 & 0.9746 & 0.9747 \\
MLP (128) & 0.9798 & 0.9797 & 0.9797 \\
MLP (256) & 0.9827 & 0.9824 & 0.9828 \\
MLP (512) & 0.9817 & 0.9816 & 0.9815 \\
\midrule
Soft Dominance (64) & 0.8959 & 0.8950 & 0.8967 \\
Soft Dominance (128) & 0.9061 & 0.9046 & 0.9119 \\
Soft Dominance (256) & 0.8930 & 0.8930 & 0.8957 \\
Soft Dominance (512) & 0.9055 & 0.9037 & 0.9068 \\
\midrule
Soft Dominance Ann. (64) & 0.9016 & 0.9006 & 0.9015 \\
Soft Dominance Ann. (128) & 0.9113 & 0.9103 & 0.9109 \\
Soft Dominance Ann. (256) & 0.9124 & 0.9116 & 0.9136 \\
Soft Dominance Ann. (512) & 0.9173 & 0.9162 & 0.9181 \\
\bottomrule
\end{tabular}
\end{table}

Each configuration was evaluated in a single training run. Table~\ref{table:results} reports descriptive observations rather than estimates of expected performance across random initializations. No standard deviations or confidence intervals across seeds are available, and small differences should not be interpreted as reliable configuration rankings. The highest observed MLP accuracy is $0.9827$ for 256 hidden units, with recall $0.9824$ and precision $0.9828$.

Without annealing, the highest observed Soft Dominance accuracy is $0.9061$ for 128 hidden units. The 512-unit result is $0.9055$, only $0.0006$ (0.06 percentage points) lower; this difference does not establish superiority of either width. The non-annealed accuracies are not monotonic in width in these runs, but variability across seeds has not been assessed.

Annealed runs yield higher accuracy and recall than the corresponding non-annealed runs at each evaluated width. The largest observed accuracy difference occurs at 256 units, from $0.8930$ to $0.9124$. The annealed 512-unit run reaches accuracy $0.9173$, recall $0.9162$, and precision $0.9181$. Its accuracy exceeds the highest non-annealed value by $0.0112$ (1.12 percentage points), but this comparison also involves different widths. Repeated-seed comparisons at matched widths are required to assess whether the observed annealing differences are robust.

The highest observed annealed Soft Dominance accuracy remains $0.0654$ (6.54 percentage points) below the highest MLP accuracy. These preliminary results demonstrate trainability in the reported MNIST experiment, but do not establish a reliable annealing benefit, a predictive advantage over the affine--ReLU baseline, or effectiveness on other datasets.

The following qualitative analysis uses the annealed 512-unit Soft Dominance run, which has the highest observed Soft Dominance accuracy in Table~\ref{table:results}; it is not established as the optimal configuration.

\begin{figure}[t]
\centering
\includegraphics[width=0.8\linewidth]{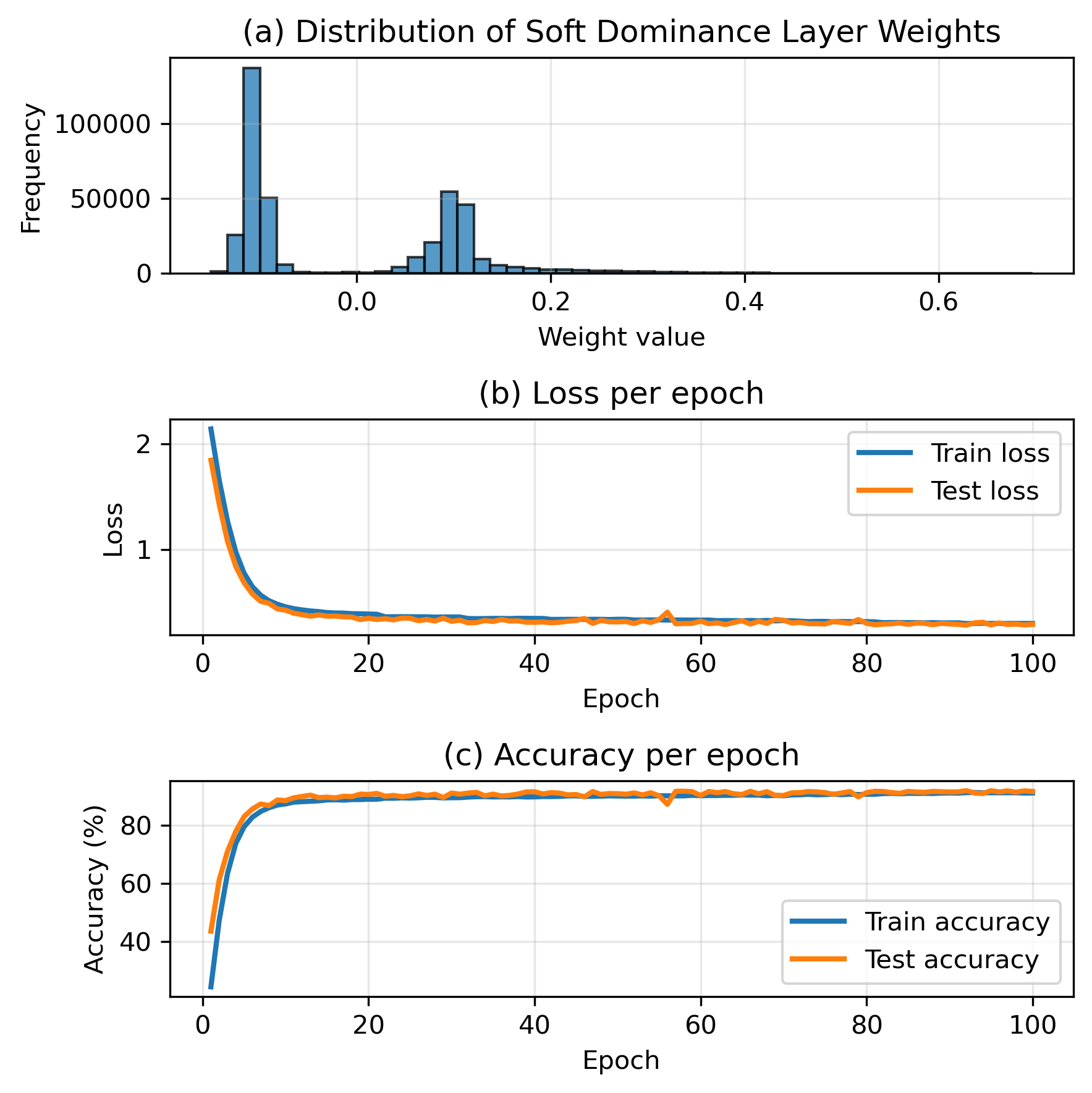}
\caption{Training diagnostics for the annealed 512-unit Soft Dominance model: (a) distribution of the learned threshold values, (b) training and test loss by epoch, and (c) training and test accuracy by epoch.}
\label{fig:soft-dominance-summary}
\end{figure}

Figure~\ref{fig:soft-dominance-summary} shows that most of the reduction in loss and increase in accuracy occur during the early epochs, followed by a slower stabilization. The training and test curves remain close throughout optimization, indicating a limited generalization gap in this run. The learned thresholds exhibit a structured, nonuniform distribution rather than collapsing to a single value.

\begin{figure}[t]
\centering
\begin{subfigure}{\linewidth}
    \centering
    \includegraphics[width=0.8\linewidth]{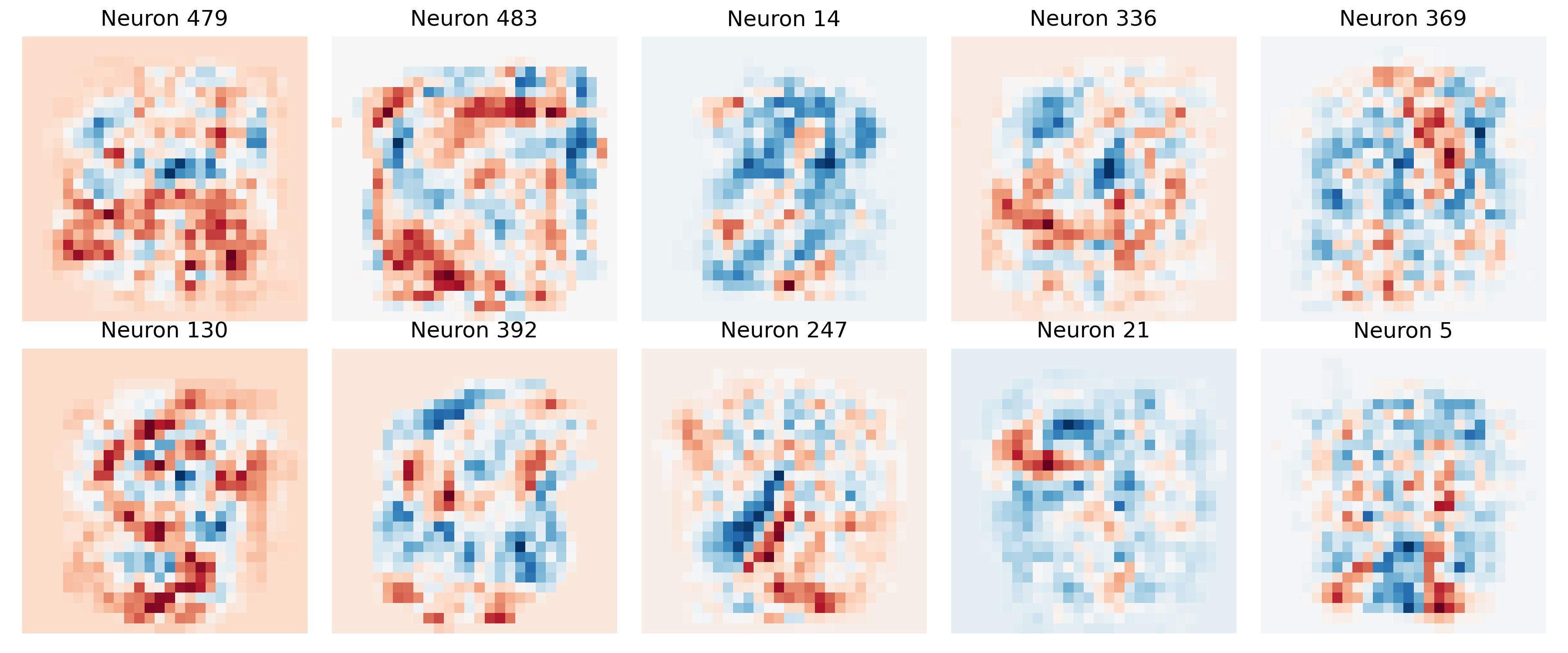}
    \caption{Learned input weights of ten selected neurons from the MLP with 512 hidden units.}
    \label{fig:mnist-mlp-neurons}
\end{subfigure}

\vspace{0.5em}

\begin{subfigure}{\linewidth}
    \centering
    \includegraphics[width=0.8\linewidth]{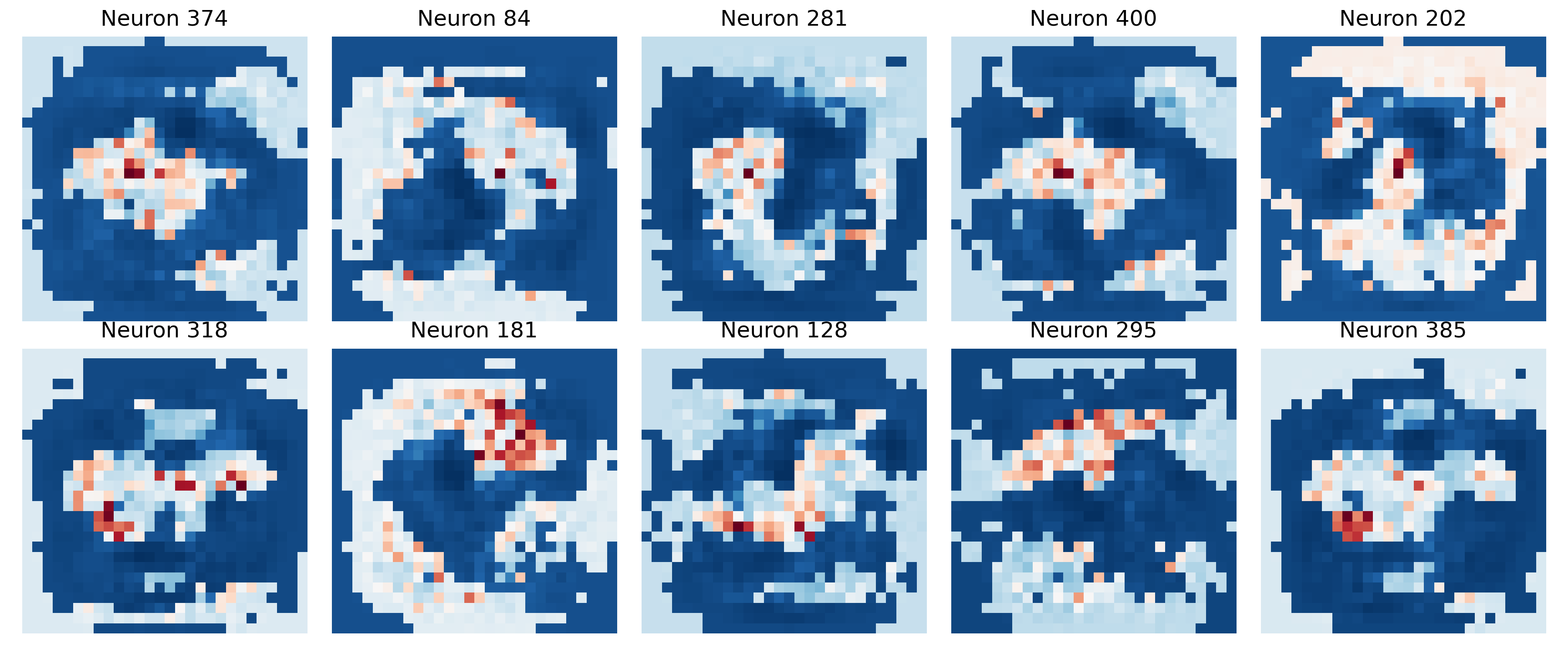}
    \caption{Learned threshold vectors of ten selected neurons from the annealed Soft Dominance model with 512 hidden units.}
    \label{fig:mnist-soft-dominance-neurons}
\end{subfigure}
\caption{Comparison of learned hidden-unit parameters. Each 784-dimensional vector is reshaped into the $28\times28$ spatial arrangement of an MNIST image: (a) weights from the conventional MLP and (b) thresholds from the proposed annealed Soft Dominance model.}
\label{fig:mnist-dominance-neurons}
\end{figure}

Figure~\ref{fig:mnist-dominance-neurons} provides a qualitative comparison of the parameters learned by the two architectures. The MLP weights in panel~(a) contain localized positive and negative variations, whereas the Soft Dominance thresholds in panel~(b) tend to exhibit broader spatial regions and nonuniform structure. The comparison should be interpreted cautiously because the two parameter types have different mathematical roles: MLP weights are projection coefficients, while Soft Dominance parameters are coordinate-wise thresholds. The visualization therefore provides evidence that the thresholds do not collapse to a trivial constant configuration, but it should not be interpreted as demonstrating class prototypes or semantic feature detectors.

\begin{figure}[t]
\centering
\includegraphics[width=0.9\linewidth]{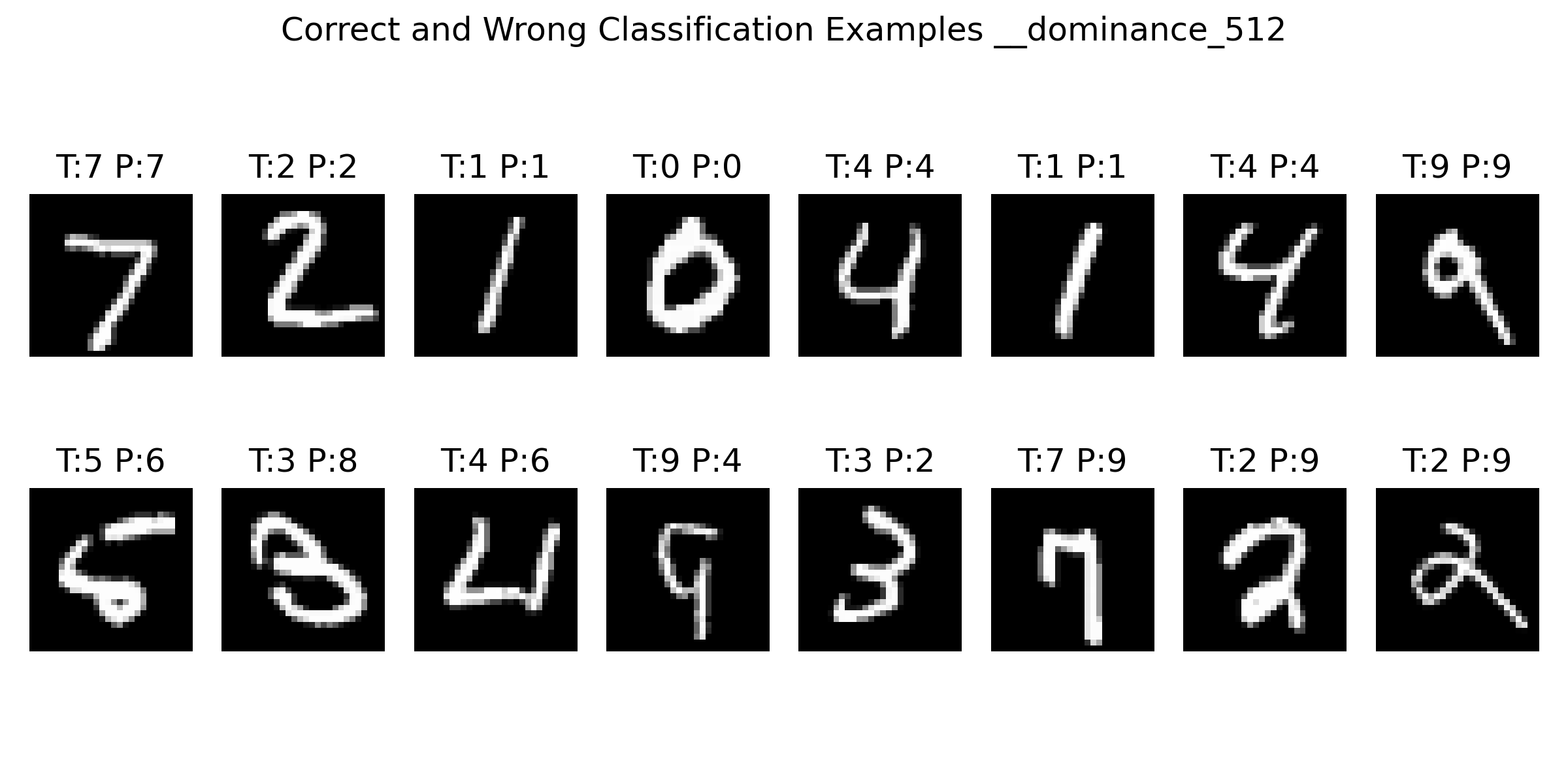}
\caption{Representative predictions from the annealed 512-unit Soft Dominance model. The first row contains correctly classified images, and the second row contains misclassified images. T denotes the true class and P the predicted class.}
\label{fig:mnist-classification-examples}
\end{figure}

Figure~\ref{fig:mnist-classification-examples} illustrates both successful predictions and typical failure cases. Several misclassified digits contain ambiguous or atypical strokes that visually resemble the predicted classes. Because the table reports individual experimental results without measures of variability, further runs with different random seeds are required to assess the statistical stability of the observed differences.

\section{Discussion}

The main result of this study is not a gain in MNIST accuracy, but the demonstration that a neural hidden layer can be constructed around a computational primitive that is qualitatively different from an affine projection. The Soft Dominance Layer learns coordinate-wise thresholds and aggregates smooth comparison outcomes, giving each parameter $W_{ij}$ a direct role as a reference value rather than as a projection coefficient.

The reported MNIST runs provide preliminary evidence that this operator can be trained with standard back-propagation. Annealed runs have higher observed accuracy at each evaluated width, making comparison sharpness a candidate for further study. However, single runs cannot establish the robustness of these differences. The highest observed annealed accuracy is $0.9173$, compared with $0.9827$ for the MLP. These observations support a limited proof of concept, not a claim of classifier superiority or reliable improvement from annealing.

One possible contributor to the observed performance gap is the structure of the current operator. Each Soft Dominance unit is additive across coordinates and individually non-increasing in each input component, restricting interactions within that unit. This explanation remains a hypothesis: the present experiments do not separate representation limits from optimization effects or random-seed variability.

The parameter visualizations reinforce the distinction between the two computational primitives. The learned Soft Dominance thresholds exhibit nonuniform spatial organization rather than collapsing to identical values, but this observation should be regarded as qualitative evidence of structured learning rather than a strong interpretability claim. A dedicated interpretability study would require quantitative analyses of threshold stability, sensitivity, and correspondence with input features.

Several limitations remain. The evaluation uses only MNIST and reports individual runs rather than statistics over multiple random seeds. It therefore does not establish stable width rankings, reproducible annealing gains, or practical relevance beyond MNIST. Repeated-seed evaluations reporting mean and standard deviation, together with experiments on additional datasets, are necessary. The comparison is width-matched but not parameter- or compute-matched, and no runtime or energy measurements are reported. Moreover, the test split is monitored across epochs; future experiments should use a validation split for development and reserve the test set for final evaluation. The current paper makes no claim of computational efficiency or general predictive advantage.

The natural next step is to explore the design space opened by the comparison-based formulation: alternative aggregation rules, learned or feature-dependent comparison directions, combinations with affine residual paths, theoretical approximation properties, and hardware-friendly approximations to the smooth comparison. Those extensions are intentionally deferred to a more complete study. The narrower contribution here is to show that moving beyond the affine transformation is technically viable and may lead to neural layers with different geometries and inductive biases.

\section{Conclusion}

This paper introduced the Soft Dominance Layer as a simple example of neural computation beyond the conventional affine transformation. For each learned reference vector $\bm W_i$, the layer computes
\begin{equation}
    \sum_j \sigma\!\left(\alpha(W_{ij}-x_j)\right),
\end{equation}
which provides a differentiable approximation to the number of coordinate-wise inequalities $x_j<W_{ij}$ that are satisfied. A normalized version yields a soft dominance score in $[0,1]$.

The main conceptual distinction is that a conventional neuron represents a projection, whereas a Soft Dominance unit represents learned coordinate-wise thresholds followed by aggregation. The single-run MNIST experiments provide preliminary evidence of end-to-end trainability. The highest observed accuracy is $0.9061$ without annealing and $0.9173$ with annealing, compared with $0.9827$ for the MLP. These values are descriptive and do not establish reliable configuration rankings or a statistically supported annealing benefit. The current additive formulation is an initial construction, not a replacement for affine layers.

Repeated-seed experiments and evaluations beyond MNIST are required before drawing conclusions about robustness or practical relevance. Further work should also examine expressive power, optimization, and computational cost. The contribution of this preliminary study is limited to formulating and demonstrating a trainable comparison-based layer in a proof-of-concept setting.

\paragraph{AI-assistance disclosure.}
This manuscript was prepared with the assistance of an artificial-intelligence-based language tool, which was used to improve its readability, grammar, and linguistic clarity. The author reviewed and approved all revisions and remains fully responsible for the scientific content, interpretations, and conclusions presented in the paper.

\appendix

\section{Mathematical Derivations}

For fixed input $\bm x$ and sharpness $\alpha$, consider the response
\begin{equation}
p_{ij}
    =
    \sigma\!\left(
        \alpha \left[\tanh(\Theta_{ij})-x_j\right]
    \right).
\end{equation}

Using $\sigma^\prime(u)=\sigma(u)\bigl(1-\sigma(u)\bigr)$ and $\tanh^\prime(v)=\operatorname{sech}^2(v)$, the chain rule gives the partial derivative with respect to $\Theta_{ij}$:
\begin{equation}
\frac{\partial p_{ij}}{\partial \Theta_{ij}} = \alpha \, \operatorname{sech}^2(\Theta_{ij}) \, p_{ij} (1 - p_{ij}).
\end{equation}

\section{PyTorch Implementation}

A direct implementation of the normalized or unnormalized Soft Dominance Layer using PyTorch~\cite{paszke2019pytorch} is shown below.

When enabled, the bias is initialized from a zero tensor with one entry per output unit. This tensor is passed explicitly to \texttt{nn.Parameter}; \texttt{trainable\_bias} controls whether it receives gradients.

\noindent\begin{minipage}{\linewidth}
\begin{lstlisting}[basicstyle=\ttfamily\fontsize{9}{10}\selectfont]
import torch
import torch.nn as nn

class SoftDominanceLayer(nn.Module):
    def __init__(self,
                 in_features,
                 out_features,
                 bias=True,
                 trainable_weight=True,
                 trainable_bias=True,
                 alpha=10.0,
                 normalize=False,
                ):
        super().__init__()

        self.in_features  = in_features
        self.out_features = out_features
        self.alpha        = alpha
        self.normalize    = normalize

        theta = torch.empty(out_features, in_features)
        nn.init.uniform_(theta, -0.5, 0.5)

        self.theta = nn.Parameter(theta,
                        requires_grad=trainable_weight)
        if bias:
            bias_tensor = torch.zeros(out_features)
            self.bias = nn.Parameter( bias_tensor,
                        requires_grad=trainable_bias)
        else:
            self.register_parameter("bias", None)

    @property
    def weight(self):
        # Constrained to (-1, 1)
        return torch.tanh(self.theta)

    def forward(self, x):
        # x: [batch, in_features]
        # W: [out_features, in_features]
        W = self.weight
        
        # [batch, out_features, in_features]
        diff = W.unsqueeze(0) - x.unsqueeze(1)
        
        # Soft coordinate-wise comparisons
        p = torch.sigmoid(self.alpha * diff)
        
        # Soft count of dominated coordinates
        y = p.sum(dim=-1)

        if self.normalize:
            y = y / self.in_features

        if self.bias is not None:
            y = y + self.bias

        return y
\end{lstlisting}
\end{minipage}

With epochs indexed from one, the following code is executed after completing each training epoch. The first update occurs after epoch 20, followed by updates after epochs 30, 40, and so on, consistently with Eq.~\eqref{eq:alpha-annealing}. The updated value is used in the following epoch.

\begin{lstlisting}
if epoch >= 20 and (epoch % 10) == 0:
    layer.alpha *= 1.3
\end{lstlisting}

\renewcommand{\refname}{References}

\end{document}